\documentclass{article}
\usepackage{ijcai23}

\usepackage{times}
\usepackage{soul}
\usepackage{url}
\usepackage[hidelinks]{hyperref}
\usepackage[utf8]{inputenc}
\usepackage[small]{caption}
\usepackage{graphicx}
\usepackage{amsmath}
\usepackage{amsthm}
\usepackage{booktabs}
\usepackage{algorithm}
\usepackage{algorithmic}
\usepackage[switch]{lineno}
\usepackage{amsfonts} 

\title{FedET: A Communication-Efficient Federated Class-Incremental Learning Framework Based on Enhanced Transformer}

\author{
Chenghao Liu$^{1,2}$
\and
Xiaoyang Qu$^1$\and
Jianzong Wang$^{1}$\footnote{Corresponding author: Jianzong Wang, jzwang@188.com} \And
Jing Xiao$^1$
\affiliations
$^1$Ping An Technology (Shenzhen) Co., Ltd., Shenzhen, China\\
$^2$The Shenzhen International Graduate School, Tsinghua University, China
\emails
liucheng21@mails.tsinghua.com.cn,
\{quxiaoyang343, wangjianzong347, xiaojing661\}@pingan.com.cn
}

\begin{document}

\maketitle

\begin{abstract}
Federated Learning (FL) has been widely concerned for it enables decentralized learning while ensuring data privacy. However, most existing methods unrealistically assume that the classes encountered by local clients are fixed over time. After learning new classes, this assumption will make the model's catastrophic forgetting of old classes significantly severe. Moreover, due to the limitation of communication cost, it is challenging to use large-scale models in FL, which will affect the prediction accuracy. To address these challenges, we propose a novel framework, \textit{Federated Enhanced Transformer} (\textbf{FedET}), which simultaneously achieves high accuracy and low communication cost. Specifically, FedET uses Enhancer, a tiny module, to absorb and communicate new knowledge, and applies pre-trained Transformers combined with different Enhancers to ensure high precision on various tasks. To address local forgetting caused by new classes of new tasks and global forgetting brought by non-i.i.d (non-independent and identically distributed) class imbalance across different local clients, we proposed an Enhancer distillation method to modify the imbalance between old and new knowledge and repair the non-i.i.d. problem. Experimental results demonstrate that FedET's average accuracy on representative benchmark datasets is 14.1\% higher than the state-of-the-art method, while FedET saves 90\% of the communication cost compared to the previous method.
\end{abstract}

\section{Introduction}\label{S1}

Federated learning (FL) enables each participating local client to benefit from other clients' data while ensuring client's data does not leave the local \cite{yang2019federated,DBLP:journals/corr/abs-2302-00903}. On the premise of ensuring the data privacy of all clients, the problem of data silos has been successfully solved \cite{hong2021federated,qu2020quantization}. However, most existing FL methods are modelled in static scenarios, meaning the models' classes are preset and fixed, which undoubtedly reduces the model's generality. Therefore, Federated Class-Incremental Learning (FCIL) is proposed. FCIL solves the problem that FL needs to retrain the entire model when meeting the new classes, saving time and computing costs. For FCIL, how to deal with catastrophic forgetting, seek the plasticity-stability balance of the model and ensure the cooperation of multiple parties are the keys to the problem.

To date, less work has been done on FCIL studies. The research conducted by \cite{hendryx2021federated} focuses on global IL by facilitating knowledge sharing among diverse clients. However, the author overlooks the non-i.i.d distribution of classes across these distinct clients. The paper \cite{dong2022federated} draw on the regularization methods used in Incremental Learning (IL) and proposes two loss functions. One for addressing the issue of forgetting old classes after IL, and the other is concentrate on the global forgetting caused by the non-i.i.d (non-independent and identically distributed) distribution of classes among different clients. However, this method needs a proxy server to achieve its best performance, leading to high communication costs and some privacy issues. To raise the accuracy of the model in FCIL settings, a natural idea is to choose a more powerful backbone model. We note that there is still no work to apply transformers to FCIL, and the biggest obstacle is that the communication cost is extremely high and cannot be reduced, which makes this application unrealistic. From another perspective, the accuracy and application scope will be significantly improved if we solve the communication and non-i.i.d. problem of class distribution between different clients.

Driven by these ideas, we propose a new \underline{Fed}erated \underline{E}nhanced \underline{T}ransformer (\textbf{FedET}) framework. Compared with other existing FCIL methods, FedET has better prediction performance, lower communication volume, and more universality. It has achieved excellent performance in both Computer Vision (CV) and Natural Language Process (NLP) fields, also it is more efficient when dealing with catastrophic forgetting. FedET consists of four main components: Pre-trained Transformer Blocks, Enhancer Select Module, Enhancer Pool and Sample Memory Module (only the local clients have the Sample Memory Module). FedET first divides the entire label space into multiple domains, each with its corresponding Enhancer Group. When new classes need to learn, Enhancer Select Module will determine which domain the new classes belong to and train a temporary Enhancer Group. The new Enhancer Group is obtained by performing distillation between the temporary Group and the corresponding old one. In this way, not only can the local clients have the capability of IL, but large-scale models (such as MAE \cite{He_2022_CVPR}) can also be used. At the same time, because only the parameters of the chosen Enhancer Group need to be updated, the communication cost is significantly reduced. 

We make the following contributions:

\begin{itemize}
    \item We introduce FedET in order to address the FCIL problem, which mitigates the issue of catastrophic forgetting in both local and global models and effectively reduces communication overhead. According to our knowledge, it is the first effort to explore the FCIL problem in a large-scale model.
    \item We propose the first FCIL framework used in both CV and NLP fields. Using different transformers as backbones, FedET can handle problems in multiple fields. Compared with baseline models, FedET improves the average accuracy of image classification by 3\% and text classification by 1.6\%.
    \item We develop a new loss to handle global catastrophic forgetting named entropy-aware multiple distillation. This is the first time an FCIL model incorporating entropy as a factor when setting the loss function.
    \item We combine the IL problem of text classification with FL for the first time. By discussing the experimental design method and baseline selection, we think it is a new challenge for both NLP and FCIL fields.
\end{itemize}

\begin{figure}[t]
\centering
\includegraphics[width=0.95\columnwidth]{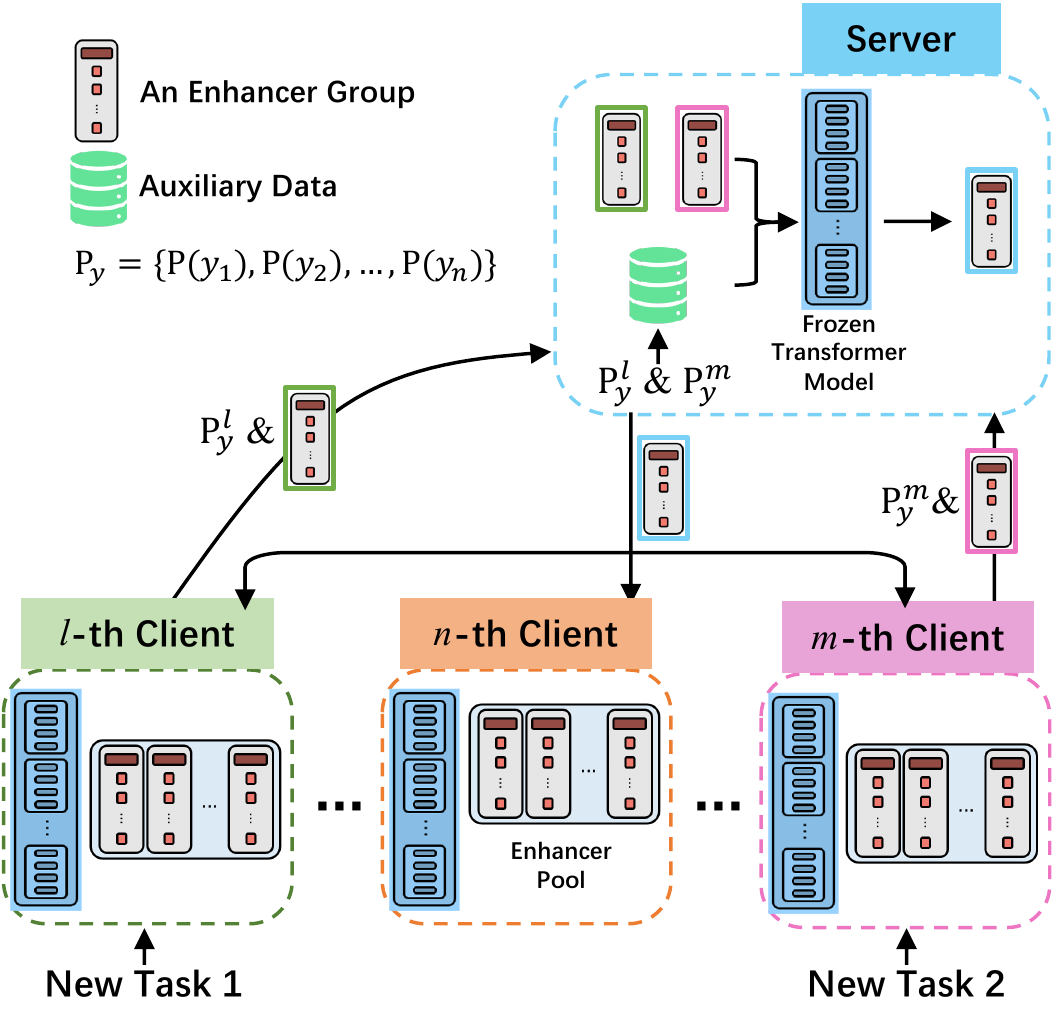}
\caption{Simple FedET scenario when performing incremental learning. Local clients upload the weight of the selected Enhancer Group and the label distribution ($\text{P}_y$) of their private training data to the server after updating the group with new tasks. Then the server uses $\text{P}_y$ to construct auxiliary data, use auxiliary data to distil upload groups, and send the updated group to all local clients.}
\label{fig_overall}
\end{figure}

\section{Preliminary}
In standard IL \cite{rebuffi2017icarl,simon2021learning,shmelkov2017incremental}, the streaming task sequence is defined by $\mathcal{T}=\{\mathcal{T}^t\}^T_{t=1}$, in which $T$ represents the task order, the first $t$ tasks $\mathcal{T}^t=\{\mathbf{x}^t_i,\mathbf{y}^t_i\}^{N^t}_{i=1}$ contains $N^t$ pairs the sample $\mathbf{x}^t_i$ and the corresponding one-hot encoded label $\mathbf{y}^t_i\in\mathcal{Y}^t$. $\mathcal{Y}^t$ represents the label space of the $t$-th task, which includes the new classes $M^t=\bigcup^B_{b=1}m^t_b$ that have not appeared in the previous $t-1$ tasks, and $B$ represents the number of new classes. At this time, the set of all classes that the model can judge is $M^A =\bigcup^t_{i=1}M^i$. Inspired by \cite{ermis2022memory,liu2020mnemonics}, based on the unique architecture of FL, we construct a Sample Memory Module $\mathcal{S}$ located on every local client to store $\frac{|\mathcal{S}|}{M^A}$ exemplars of each class at local, and it satisfies $\frac{|\mathcal{S}|}{M^A}\ll\frac{N^t}{M^t}$.

For FCIL, we give the initial setting under the FL framework \cite{yoon2021federated}: we set $K$ local clients $\mathcal{C}=\{\mathcal{C }_k\}^K_{k=1}$ and a global server $\mathcal{C}_G$, the model structures on all clients and server are the same, from the perspective of parameters, including the frozen parameter $\Phi$ (that is, the parameters of the selected pre-trained backbone model) and the variable parameter $\theta$. When $a$ clients ($a<K$) send applications to server for Class-Incremental Learning (CIL), they will access the $t$-th task, updated $\theta$, and select some samples $\{\mathbf{x}^ t_i,\mathbf{y}^t_i\}$ put into Sample Memory Module. 

\begin{figure*}[t]
\centering
\includegraphics[width=0.95\textwidth]{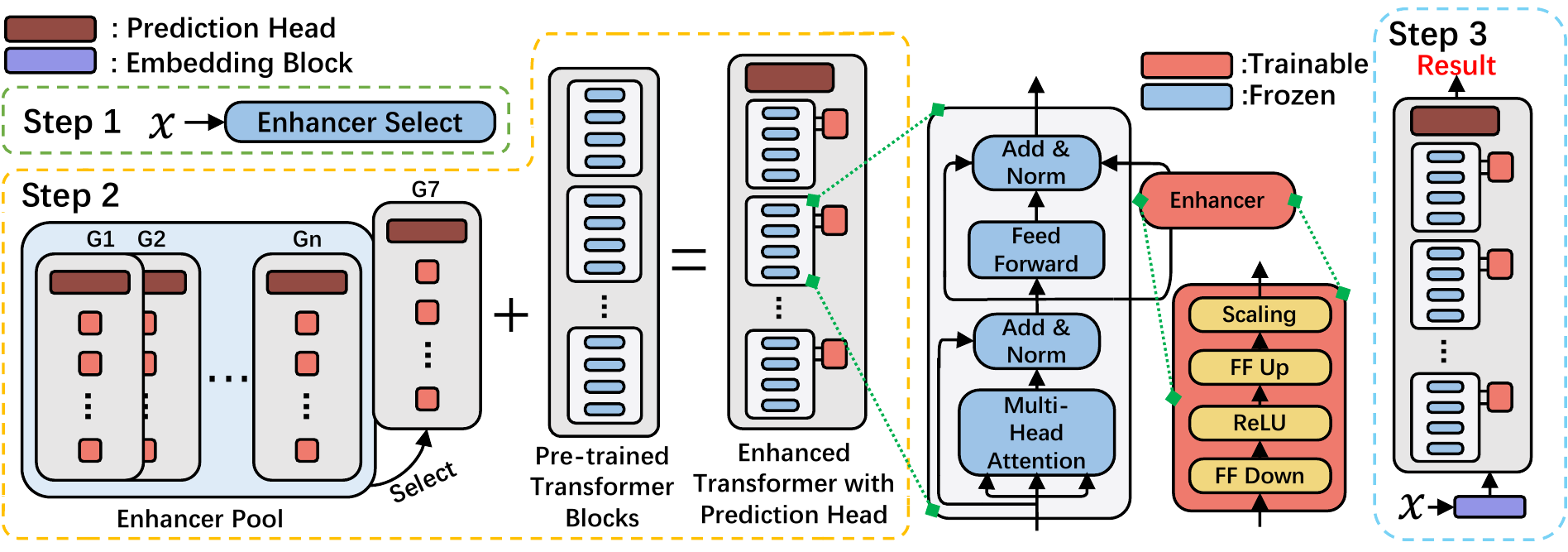} 
\caption{The workflow of making predictions by the local model in FedET. The input will first be processed by the Enhancer Select Module to decide the right Enhancer Group. Then this Group will insert into the pre-trained backbone and develop the prediction model. It should be noted that only the parameters of Enhancers are trainable, and the seventh group is used as an example in the figure.}
\label{fig_Local}
\end{figure*}

\section{Methodology}

While CIL and FCIL share similarities, the key distinction between them is that FCIL involves tackling two types of forgetting: local and global. FedET addresses local forgetting through a dual distillation loss and mitigates global forgetting through auxiliary data construction and an entropy-aware multiple distillation loss. Figure \ref{fig_overall} shows the general outline of the FedET approach.

\subsection{Solution of Local Forgetting}
In FedET, a local model mainly includes four parts: Pre-trained Transformer Blocks, Enhancer Select Module, Enhancer Pool and Sample Memory Module. We show the local model's specific structure and predicting process in Figure \ref{fig_Local}.

\subsubsection{Enhancer Pool and Enhancer Group}
Enhancer is the core of FedET, so it is introduced here first. An Enhancer Group contains some Enhancers and a prediction head. And the number of Enhancers is decided by the frozen Pre-trained Transformer Blocks. The mainstream methods of IL fall into three categories \cite{de2021continual}: playback, regularization, and parameter isolation. In FedET, we use a combination of three approaches: for each client, we set up an Enhancer Pool, which contains multiple Enhancer Groups $\mathcal{H}=\{\mathcal{H}^j\}^J_{j=1}$, each group is dedicated to being proficient in part of all existing classes. That is, for an Enhancer Group, the class it is responsible for is $M^{\mathcal{H}^j}$, and $\bigcup^J_{j=1}M^{\mathcal{H}^j}= M^A$. Setting the parameters of an Enhancer Group and a frozen Pre-trained transformer model as $\theta$ and $\Phi$ respectively, An Enhancer Group includes many Enhancers $\theta_E$ and a prediction head $\theta_H$. During training, only $\theta$ is modified, $\Phi$ is still frozen. Thereby greatly reducing the number of parameters that need to be adjusted without dropping accuracy. An Enhancer is comprised of three components: a down-projection with $W_{\text{down}} \in \mathbb{R}^{n\times m}$, an activation function $f(\cdot)$, and an up-projection with $W_{\text{up}} \in \mathbb{R}^{m\times n}$. Since the encoder structures of transformers are almost the same, after completing FedET's experiments in the NLP and CV fields, we believe this framework can be used for most of the currently known transformers. 

Why choose "Enhancer + Freeze the backbone model" instead of Freeze the underlying encoder to adjust the upper encoder? We draw two perspectives from experiments and the literature \cite{ruckle2020adapterdrop}. First, Enhancer is freely pluggable, and its internal structure can keep the input of the original encoder, so it can retain the maximum amount of the knowledge that the backbone model has learned during the pre-training stage. Meanwhile, through the design of the Enhancer bottleneck structure and the freezing of the pre-training model, the entire model can learn the downstream tasks better, while the number of parameters that need to be adjusted is significantly reduced. Second, we note that direct fine-tuning can easily lead to overfitting during training on downstream tasks, whereas inserting the Enhancer module performs much better. Although it can be compensated by carefully tuning hyperparameters such as learning rate and batch size, it is undoubtedly time-consuming and labour-intensive.
 
\subsubsection{Enhancer Select Module and Sample Memory Module}
After a sample is preprocessed, it will be input to the Enhancer Select Module $G_s(x)$. The Enhancer Select Module is a pre-trained frozen classifier. The function of this module is to select a suitable Enhancer Group to handle the input sample. The output of this classifier tells FedET which group is the right group to call up. In $t$-th task $\mathcal{T}^t$, the Enhancer Select Module will first select an Enhancer Group (e.g. $j$-th) $\mathcal{H}^j$ according to the judgement that the new class $M^t$ is the most similar to the class $M^{\mathcal{H}^j}$, then $\mathcal{H}^j$ will participate in distillation as $\mathcal{H}^j_{\text{old}}$. There will be a randomly initialized temporary Enhancer Group $\mathcal{H}^{t}$ aiming to study $M^t$. After the study is completed, $\mathcal{H}^{t}$ will perform distillation with $\mathcal{H}^j_{\text{old}}$ to obtain a new Enhancer Group $\mathcal{H}^j_{\text{new}}$ which covers $\mathcal{H}^j_{\text{old}}$, and the specialized class of the group change to  $M^{\mathcal{H}^j_{\text{new}}}=M ^{\mathcal{H}^j}\cup M^t$. The judgment methods of the Enhancer Select Module corresponding to different task fields are also different. For CV fields, we design the Enhancer Select Module as an EfficientNet \cite{tan2019efficientnet}, and for NLP fields, we use text-RCNN \cite{lai2015recurrent}. During the distillation process, the required old class samples $\mathcal{S}_{M^{\mathcal{H}^j_{\text{old}}}}$ are provided by the Sample Memory Module. After the new class is learned, this module will also store one typical sample of each new type $\mathcal{S}_{M^t}$, get new $\mathcal{S}_{M^{\mathcal{H}^j_{\text{new}}}}=\mathcal{S}_ {M^{\mathcal{H}^j}}\cup\mathcal{S}_{M^t}$, to ensure that the subsequent distillation can proceed smoothly.

 After selection, the parameters of the chosen Enhancer Group $\theta^{\text{old}}$ are linked with the frozen pre-trained model parameters $\Phi$. The $\theta^{\text{old}}$ contains two parts: $\theta_E^{\text{old}}$, the parameters of the Enhancers, and $\theta_H^{\text{old}}$, the parameters of the prediction head.

\subsubsection{Distillation of Enhancers}
For a new task $\mathcal{T}^t$, the new classes need to learn is $M^t$. After the temporary Enhancer Group $\mathcal{H}^{t}$ finish studying, it will be linked with the frozen transformer model as the temporary model $f_{t}$, which contains $\Phi$ and $\theta^t$. The frozen transformer model with $\mathcal{H}^{\text{old}}$ is called the old model $f_{\text{old}}$. We use the following objective to distill $f_{\text{old}}$ and $f_{t}$:
\begin{equation}
    f_{\text{new}}(x;\theta^{\text{new}},\Phi)=\left\{\begin{array}{lc}
         f_{\text{old}}(x;\theta^{\text{old}},\Phi)[i] & 1\leq i \leq m\\
         \\
         f_{t}(x;\theta^t,\Phi)[i]& m< i\leq c
    \end{array}\right.
\end{equation}
we set 
\begin{equation}
    c = m+n
\end{equation}
where $m$ and $n$ is the number of $M^{\text{old}}$ and $M^t$ respectively. To ensure that the consolidated model's output $f(x;\theta^{\text{new}},\Phi)$ approximates the combination of outputs from $f_{t}$ and $f_{\text{old}}$, we utilize the output of $f_{t}$ and $f_{\text{old}}$ as supervisory signals during the joint training of the consolidated parameters $\theta^{\text{new}}$.

To achieve this goal, we employ the double distillation loss proposed by \cite{zhang2020class} to train $f_{\text{new}}$. The distillation process is as follows: $f_{t}$ and $f_{\text{old}}$ are frozen, and run a feed-forward pass with every sample in training set to collect the \textbf{logits} of $f_{t}$ and $f_{\text{old}}$:
$$\hat{y}_{\text{old}}=\{\hat{y}^1,\cdots,\hat{y}^m\},\quad\hat{y}_{t}=\{\hat{y}^{m+1},\cdots,\hat{y}^{m+n}\}$$
respectively, where the superscript is the class label. Then main requirements is to reduce the gap between the \textbf{logits} generated by $f_{\text{new}}$ and the \textbf{logits} generated by $f_{t}$ and $f_{\text{old}}$. Based on prior work\cite{zhang2020class}, We choose L2 loss \cite{ba2014deep} as the distance metric. Speciﬁcally, the training objective for consolidation is:
\begin{equation}\label{eq_min}
    \min\limits_{\theta^{\text{new}}}\frac{1}{|\mathcal{U}|}\sum_{x_i\in\mathcal{U}} L_{dd}(y,\dot{y})
\end{equation}
where $\mathcal{U}$ denotes the training samples from Sample Memory Module used for distillation. And $L_{dd}$ is the double distillation loss:
\begin{equation}\label{eq_ld}
    L_{dd}(y,\dot{y})=\frac{1}{m+n}\sum^{m+n}_{i=1}(y^i-\dot{y}^i)^2
\end{equation}
in which $y^i$ are the \textbf{logits} produced by $f_{\text{new}}$ for the $t$-th task, and
\begin{equation}\label{eq_y}
    \dot{y}^i=\left\{\begin{array}{lc}
         \hat{y}_{i}-\frac{1}{m}\sum\limits^m\limits_{j=1}\hat{y}_{j}& 1\leq i\leq m \\
         \\
         \hat{y}_{i}-\frac{1}{n}\sum\limits^{m+n}\limits_{j=m+1}\hat{y}_{j}& m < i\leq c
    \end{array}\right.
\end{equation}
where $\hat{y}$ is the concatenation of $\hat{y}_{\text{old}}$ and $\hat{y}_{\text{new}}$. After the consolidation, the Enhancers parameters $\theta^{\text{new}}$ are used for $f_{\text{old}}$ in the next round. The pseudo code for local forgetting solution is shown in Algorithm \ref{alg1}.

\begin{algorithm}[h] 
	\renewcommand{\algorithmicrequire}{\textbf{Input:}}
	\renewcommand{\algorithmicensure}{\textbf{Output:}}
	\caption{Local\_ICL} 
	\label{alg1} 
	\begin{algorithmic}[1]
		\REQUIRE{Enhancer Select Module $G_s(x)$}
        \REQUIRE{Enhancer Pool $\mathcal{H}=\{\mathcal{H}^j\}^J_{j=1}$}
		\REQUIRE{Sample Memory Module $\mathcal{S}$}
        \REQUIRE{Parameters of pre-trained transformer model $\Phi$}
        \REQUIRE{$t$-th task data $\mathcal{T}^t=\{\mathbf{x}^t_i,\mathbf{y}^t_i\}^{N^t}_{i=1}$}
        \FOR{$i=1\to N^t$}
		\STATE {Group number $j \leftarrow G_s(\mathbf{x}^t_i)$}
        \STATE {Put $\{\mathbf{x}^t_i,\mathbf{y}^t_i\}$ with the same $j$ into a list $L_j$}
        \ENDFOR
        \FOR{every selected $j$}
        \WHILE{Non-convergence}
        \STATE {Randomly initialize temporary group $\mathcal{H}^t$ }
        \STATE {train $f_t(x;\theta^t,\Phi)$ with $L_j$}
        \STATE {Sample from $L_j$ and add to $\mathcal{S}$}
        \STATE {$f_{\text{new}}(x;\theta^{\text{new}},\Phi) = \textit{DISTILLATION}(f_t,f_{\text{old}},\mathcal{S})$}
        \STATE {$\theta^j\leftarrow\theta^{\text{new}}$}
        \STATE {Communicate $\mathcal{H}^j$ with Server to get Global best $j$-th group in this turn}
        \ENDWHILE
        \ENDFOR
        \STATE{}
        
        \textit{DISTILLATION} {$f_t,f_{\text{old}},\mathcal{S}$}
        
        \STATE{Get $\hat{y}_{\text{old}}$ from $f_{\text{old}}$ and $\mathcal{S}$}
        \STATE{Get $\hat{y}_t$ from $f_t$ and $\mathcal{S}$}
        \STATE{Compute loss function as in Eq.\ref{eq_ld} and train $f_{\text{new}}$}
        \RETURN{$f_{\text{new}}$}
    \end{algorithmic} 
\end{algorithm}

\subsection{Solution of Global Forgetting}
Global catastrophic forgetting primarily arises from the heterogeneity forgetting among local clients participating in incremental learning. Which means the non-i.i.d. class-imbalanced distributions across local clients lead to catastrophic forgetting of old classes on a global scale, further exacerbating local catastrophic forgetting. Therefore, it is necessary to solve the heterogeneity forgetting problem across clients in global perspective. To ensure precision and speed, FedET handles this problem with double distillation loss and the difference of average entropy across different clients.

\subsubsection{Distillation of Enhancers of Different Clients}
FedET changed the stereotype of having to queue up for updates and proposed a new way to update the model. The new method is more scientific, reasonable, and time-effective. When a single client uploads the new Enhancer Group obtained after distillation to the server, all the server needs to do is update the parameters of the corresponding group. When many clients upload the same Enhancer simultaneously, queuing is unscientific because only the last client's update is critical, and this is how global catastrophic forgetting happens. FedET sets a server waiting time limitation. Within a specific time, multiple schemes for an Enhancer Group will be aggregated by the server to perform global model distillation. 

For global distillation, the server will distill some Enhancer Group at same time, which means there will be $f_t^1,f_t^2,\cdots,f_t^q (q<\textit{the number of clinets})$ and a $f_{\text{old}}$ distill together. Note that the new classes are learned by all uploaded groups. Suppose the class which the distilled Enhancer Group major in is $M^t$. For every group-uploaded client, they also upload the information entropy $H(M^t)$ of $M^t$ to the server. The server uses $H(M^t)$ to judge the importance of each $f_t$, in detail, the consolidated model of global distillation is:

\begin{equation}
        f_{\text{new}}(x;\theta^{\text{new}},\Phi)=
        \left\{\begin{array}{lc}
         f_{\text{old}}(x;\theta^{\text{old}},\Phi)[i] & 1\leq i \leq m\\
         \\
         \sum\limits^q\limits_{k=1} \frac{H^k}{H_{\text{sum}}} f_t^k(x;\theta^t,\Phi)[i] & m< i\leq c
    \end{array}\right.
\end{equation}

where $H_{\text{sum}}$ is the sum of information entropy $H(M^t)$ of all uploaded clients. Noted that all output of $f$ here are \textbf{logits}, not hard-label. To get $\theta^{\text{new}}$, the entropy-aware multiple distillation loss $L_{emd}$ is:
\begin{equation}
    L_{emd}(y,\ddot{y})=\frac{1}{m+n}\sum^{m+n}_{i=1}(y^i-\ddot{y}^i)^2
\end{equation}
in which $\ddot{y}$ is:
\begin{equation}
        \ddot{y}^i=\left\{\begin{array}{lc}
         \hat{y}_{i}-\frac{1}{m}\sum\limits^m\limits_{j=1}\hat{y}_{j}& 1\leq i\leq m \\
         \\
         \hat{y}_{i}-\frac{1}{nH_{\text{sum}}}\sum\limits^{m+n}\limits_{j=m+1}\sum\limits^q\limits_{k=1} H^k\hat{y}_j^k& m < i\leq c
    \end{array}\right.
\end{equation}

Because of the nature of FL, we cannot rely solely on the sampled data to consolidate the updated Enhancers. Therefore, auxiliary data must be used. During local Enhancer distillation, we generate $\mathcal{U}$ using the Sample Memory Module, which stores one representative sample per class and utilizes data augmentation to create the dataset. For global distillation, we construct an equivalent dataset to approximate the training samples. After the local clients send the label distribution $P_y$ to the server, the server can construct the auxiliary datasets using available data of a similar domain. Notably, these auxiliary datasets are dynamically fetched and inputted in mini-batches, reducing the storage burden, and discarded after distillation is complete.

\begin{table}[t]
    \centering
    \begin{tabular}{|c|c|c|c|}
        \hline
        \textbf{Model} & \textbf{Method} &\begin{tabular}{c}\textbf{Updated}\\\textbf{Paras.}\end{tabular} &\begin{tabular}{c}\textbf{Training}\\\textbf{Time}\end{tabular}\\\hline

        BERT& Fine-tuning & $110.01\times 10^6$ & 1.92 sec \\\cline{2-4} 
                              & Enhancer & $1.76\times 10^6$ & 1.19 sec \\\hline
        ViT-Base & Fine-tuning & $75.99\times 10^6$ & 0.94 sec \\\cline{2-4} 
                              & Enhancer & $1.19\times 10^6$ & 0.59 sec \\\hline
        

    \end{tabular}
    \caption{The communication cost and computation cost difference between whether inserting Enhancer or not. Here Updated Paras. refers to the number of updated parameters}
    \label{tab:CommunicationC}
\end{table}

\subsubsection{Communication Cost Analysis}
The parameter quantity of a single Enhancer is $2mn+n+m$. For a single local model, if there are $D$ Encoder modules in one Enhancer Group, after adding a group of Enhancers, the increased parameter quantity is:
\begin{equation} \label{eq2}
    D\times(2mn+n+m)+n\times labels
\end{equation}
Other parameters are frozen except for the Enhancer Group and prediction head in the model. As shown in Table \ref{tab:CommunicationC}, the network parameters that need to be transmitted are reduced by more than 70\% compared with the various FCIL models previously proposed.

\subsubsection{Computation Cost Analysis}
The computation FLOPs for each Enhancer in the forward pass are $2 \times m\times n\times \textit{sequence length}$ (normalized to a single data sample). The overhead incurred in this way is negligible compared to the original model complexity, e.g., less than 1\% on BERT. In the meantime, since all other parameters are fixed during the training period, the computation during backpropagation is reduced by skipping the gradient that computes most of the weights. As shown in Table \ref{tab:CommunicationC}, the use of Enhancers reduces the training time by about 40\%.

\begin{table}[t]
    \centering
    \begin{tabular}{lccc}
        \hline
        \textbf{Dataset} & \textbf{Class} & \textbf{Type} & \textbf{Train / Test} \\\hline
        AGnews & 4 & News & 8000 / 2000 \\
        Yelp & 5 & Sentiment & 10000 / 2500 \\
        Amazon & 5 & Sentiment & 10000 / 2500 \\
        DBpedia & 14 & Wikipedia Article & 28000 / 7000 \\
        Yahoo & 10 & Q\&A & 20000 / 5000\\
        \hline
    \end{tabular}
    \caption{The text classification dataset we used includes statistics on various domains of classification tasks.}
    \label{tab:NLPdatasets}
\end{table}

\begin{table}[t]
    \centering
    \begin{tabular}{ll}
        \hline
        \textbf{Order} & \textbf{Task Sequence}\\\hline
        1 & AGnews$\rightarrow$Yelp$\rightarrow$Yahoo\\
        2 & Yelp$\rightarrow$Yahoo$\rightarrow$AGnews\\
        3 & Yahoo$\rightarrow$AGnews$\rightarrow$Yelp\\
        4 & AG$\rightarrow$Yelp$\rightarrow$Amazon$\rightarrow$Yahoo$\rightarrow$DBpedia\\
        5 & Yelp$\rightarrow$Yahoo$\rightarrow$Amazon$\rightarrow$DBpedia$\rightarrow$AGnews\\
        6 & DBpedia$\rightarrow$Yahoo$\rightarrow$AGnews$\rightarrow$Amazon$\rightarrow$Yelp\\
        \hline
    \end{tabular}
    \caption{Six different dataset sequences for NLP experiments}
    \label{tab:NLPdataseq}
\end{table}

\begin{table*}[htb]
\centering
\setlength{\tabcolsep}{3.3mm}
\begin{tabular}{|l|ccc|c|ccc|c|}
\hline
\textbf{Model} & \multicolumn{4}{c|}{\textbf{Length-3 Task Sequences}} & \multicolumn{4}{c|}{\textbf{Length-5 Task Sequences}} \\ \hline
\textbf{Order} & \textbf{1}    & \textbf{2}   & \textbf{3}   & \textbf{Average}   & \textbf{4}    & \textbf{5}   & \textbf{6}   & \textbf{Average}   \\ \hline
Finetune + FL & 25.79 & 36.56 & 41.01 & 34.45 & 32.37 & 32.22 & 26.44 & 30.34 \\
Replay + FL & 69.32 & 70.25 & 71.31 & 70.29 & 68.25 & 70.52 & 70.24 & 69.67   \\
Regularization + Replay + FL & 71.50 & 70.88 & 72.93 & 71.77 & 72.28 & 73.03 & 72.92 & 72.74   \\
IDBR + FL & 71.80 & 72.72 & 73.08 & 72.53 & 72.63 & 73.72 & \textbf{73.23} & 73.19 \\ 
\textbf{FedET} & \textbf{73.12}  & \textbf{73.57} & \textbf{74.28} & \textbf{73.66} & \textbf{73.83} & \textbf{74.23} & 73.18 & \textbf{73.75} \\ \hline
MTL + FL & 74.16 & 74.16 & 74.16 & 74.16 & 75.09 & 75.09 & 75.09 & 75.09 \\ \hline
\end{tabular}
\caption{Performance comparisons between FedET and other incremental text classification baseline methods}
\label{tab:NLP}
\end{table*}

\begin{table*}[t]
    \centering
    \setlength{\tabcolsep}{1.9mm}
    \begin{tabular}{|l|cccccccccc|c|c|}
    \hline
         \textbf{Methods} & \textbf{10} & \textbf{20} & \textbf{30} & \textbf{40} & \textbf{50} & \textbf{60} & \textbf{70} & \textbf{80} & \textbf{90} & \textbf{100} & \textbf{Avg.}  & \begin{tabular}{c}\textbf{Communication}\\\textbf{Cost per Task}\end{tabular}\\\hline
         iCaRL + FL & 73.5 & 61.3 & 55.7 & 45.9 & 45.0 & 39.7 & 36.7 & 33.9 & 32.2 & 31.8 & 46.5  & $10.82\times 10^6$\\
         BiC + FL & 74.3 & 63.0 & 57.7 & 51.3 & 48.3 & 46.0 & 42.7 & 37.7 & 35.3 & 34.0 & 49.0  & $10.82\times 10^6$\\
         PODNet + FL & 74.3 & 64.0 & 59.0 & 56.7 & 52.7 & 50.3 & 47.0 & 43.3 & 40.0 & 38.3 & 52.6  & $10.82\times 10^6$\\
         SS-IL + FL & 69.7 & 60.0 & 50.3 & 45.7 & 41.7 & 44.3 & 39.0 & 38.3 & 38.0 & 37.3 & 46.4  & $10.82\times 10^6$\\
         DDE + iCaRL + FL & 76.0 & 57.7 & 58.0 & 56.3 & 53.3 & 50.7 & 47.3 & 44.0 & 40.7 & 39.0 & 52.3  & $10.82\times 10^6$\\
         GLFC & 73.0 & 69.3 & 68.0 & 61.0 & 58.3 & 54.0 & 51.3 & 48.0 & 44.3 & 42.7 & 57.0  & $10.82\times 10^6$\\
         
         \hline
         \textbf{FedET($J=10$)}& \textbf{83.2} & \textbf{75.7} & \textbf{72.0} & \textbf{69.4} & \textbf{67.9} & \textbf{65.8} & \textbf{63.4} & \textbf{62.1} & \textbf{61.0} & \textbf{60.6} & \textbf{68.1}  & $1.19\times 10^6$\\\hline
    \end{tabular}
    \caption{Comparison of FedET's performance with other CV baselines in ten incremental tasks on ImageNet-Subset. During the experiment, FedET only communicates the parameter of the changed Enhancer Group, and other methods update the entire model(ResNet18).}
    \label{tab:CVImageNet}
\end{table*}

\begin{table*}[t]
    \centering
    \setlength{\tabcolsep}{1.9mm}
    \begin{tabular}{|l|cccccccccc|c|c|  }
    \hline
         \textbf{Methods} & \textbf{10} & \textbf{20} & \textbf{30} & \textbf{40} & \textbf{50} & \textbf{60} & \textbf{70} & \textbf{80} & \textbf{90} & \textbf{100} & \textbf{Avg.} & \begin{tabular}{c}\textbf{Communication}\\\textbf{Cost per Task}\end{tabular}\\\hline
         iCaRL + FL & 89.0 & 55.0 & 57.0 & 52.3 & 50.3 & 49.3 & 46.3 & 41.7 & 40.3 & 36.7 & 51.8  & $10.82\times 10^6$\\
         BiC + FL & 88.7 & 63.3 & 61.3 & 56.7 & 53.0 & 51.7 & 48.0 & 44.0 & 42.7 & 40.7 & 55.0  & $10.82\times 10^6$\\
         PODNet + FL & 89.0 & 71.3 & 69.0 & 63.3 & 59.0 & 55.3 & 50.7 & 48.7 & 45.3 & 45.0 & 59.7  & $10.82\times 10^6$\\
         SS-IL + FL & 88.3 & 66.3 & 54.0 & 54.0 & 44.7 & 54.7 & 50.0 & 47.7 & 45.3 & 44.0 & 54.9  & $10.82\times 10^6$\\
         DDE + iCaRL + FL & 88.0 & 70.0 & 67.3 & 62.0 & 57.3 & 54.7 & 50.3 & 48.3 & 45.7 & 44.3 & 58.8  & $10.82\times 10^6$\\
         GLFC & 90.0 & 82.3 & 77.0 & 72.3 & 65.0 & 66.3 & 59.7 & 56.3 & 50.3 & 50.0 & 66.9  & $10.82\times 10^6$\\

         \hline
         FedET($J=1$)&89.0& 61.0& 62.0&55.4&51.7&49.1&45.8&43.1&40.2&37.9& 53.2& $1.19\times 10^6$\\
         FedET($J=5$)&89.6&82.3&77.0&72.9&64.8&61.0&59.9&56.3&50.7&49.8& 66.4& $1.19\times 10^6$\\
         \textbf{FedET($J=10$)}& \textbf{93.1} & \textbf{84.2} & \textbf{82.4} & \textbf{79.3} & \textbf{77.4} & \textbf{74.7} & \textbf{71.4} & \textbf{68.7} & \textbf{66.5} & \textbf{66.0} & \textbf{76.4}  & \textbf{$1.19\times 10^6$}\\\hline
    \end{tabular}
    \caption{Comparison of FedET's performance with other CV baselines in ten incremental tasks on CIFAR-100. During the experiment, FedET only communicates the parameter of the changed Enhancer Group, and other methods update the entire model(ResNet18).}
    \label{tab:CVcifar}
\end{table*}

\begin{figure*}[t]
\centering
\includegraphics[width=1\textwidth]{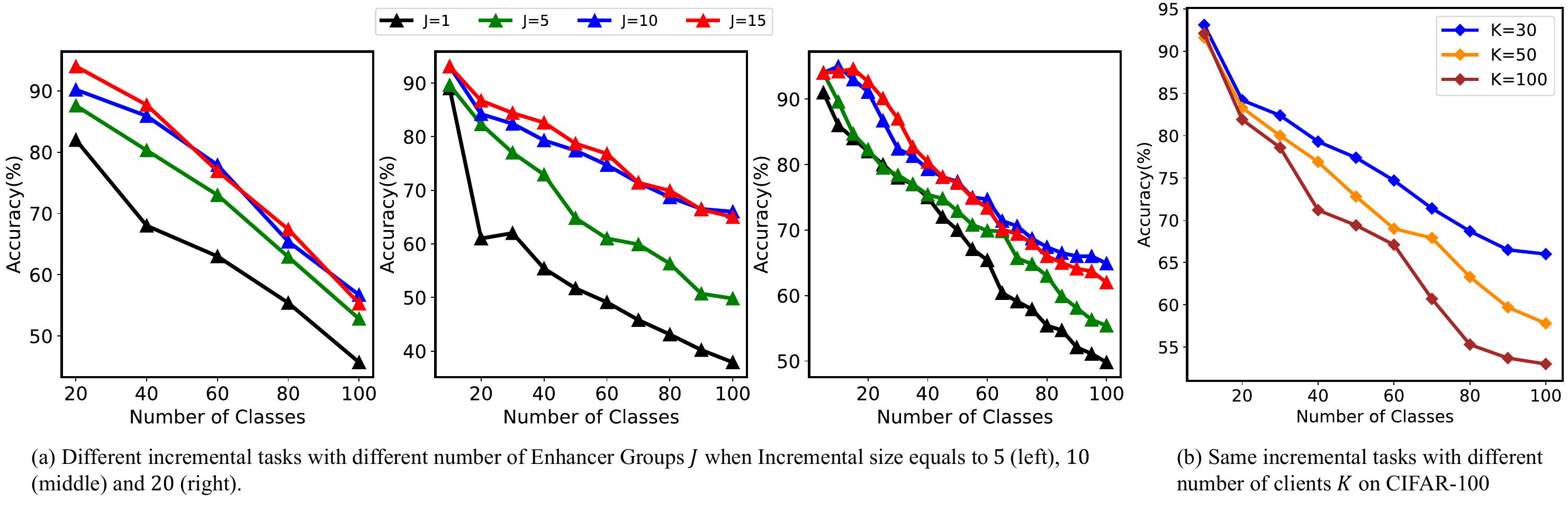} 
\caption{Ablation study}
\label{fig_ab}
\end{figure*}

\section{Experiments}
As discussed in Section \ref{S1}, since transformers are widely used in both NLP and CV fields, we test the performance of FedET on image and text classification tasks. The complete setup will be described in the following subsections.

\subsection{Natural Language Processing (NLP)}

\subsubsection{Datasets and Baselines}
Owing to the limited label space of a single dataset, we integrated five text classification datasets \cite{chen2020mixtext} to evaluate FedET. Table \ref{tab:NLPdatasets} displays the details of the dataset. Considering the domain similarity of Yelp and Amazon, we merge their label spaces for a total of 33 classes. We followed specific task sequences as outlined in Table \ref{tab:NLPdataseq} during training. To alleviate the impact of sequence length and task order on experiment results, we test task sequences of length-3 and length-5 in different orders. The first three tasks of length-3 sequences are a cyclic shift of AGnews$\rightarrow$Yelp$\rightarrow$Yahoo, which belong to three distinct domains (news classification, sentiment analysis, Q\&A classification). The remaining three task sequences of length-5 follow the experimental design proposed by \cite{de2019episodic}. During validation, the validation set comprise all classes.

Currently, there is no text classification in the FCIL field, so we choose the baseline of text classification in the Class-Incremental Learning (CIL) field and federate it to form the baseline of this experiment. We compare FedET with five baselines:
\begin{itemize}
    \item \textbf{Finetune} \cite{yogatama2019learning}\textbf{ + FL}: Only new tasks are used to fine-tune the BERT model in turn.
    \item \textbf{Replay} \cite{de2019episodic}\textbf{ + FL}: Replay some old tasks examples during new-tasks-learning to \textbf{Finetune} the model.
    \item \textbf{Regularization + Replay + FL}: On the foundation of \textbf{Replay}, add an L2 regularization term to the hidden state of the classifier following BERT.
    \item \textbf{IDBR} \cite{huang2021continual}\textbf{ + FL}: On the basis of \textbf{Regularization + Replay + FL}, replace the L2 regularization term with an information disentanglement-based regularization term.
    \item \textbf{Multi-task Learning} (MTL): Train the model with all class in one task. This approach represents an upper bound on the performance achievable through incremental learning.

\end{itemize}
\subsubsection{Implementation Details}
In this NLP experiment, we set $J=3$ Enhancer Groups and $K=10$ local clients. The prediction head for each group is a linear layer with a Softmax activation function. For sample collection (i.e. experience replay), we stored 3\% (store ratio $\gamma = 0.03$) of observed examples in the Sample Memory Module, which is used for local Enhancer distillation. We choose the pre-trained Bert-Base-Uncased from HuggingFace Transformers \cite{wolf2020transformers} as our backbone model. All experiments utilized a batch size of 16 and a maximum sequence length of 256. During training, ADAMW \cite{loshchilov2017decoupled} is used as the optimizer, with a learning rate $lr=3e^{-5}$ and a weight decay $0.01$ for all parameters. For each round of global training, three clients are randomly selected for ten epochs of local training. Selected clients are randomly given 60\% of the classes from the label space of its seen tasks.

\subsubsection{Results}
As shown in Table \ref{tab:CommunicationC} and Table \ref{tab:NLP},  we can directly see the importance of experience replay for FCIL in NLP. Moreover, the simple regularization approach based on experience replay consistently improves results across all six orders. In most cases, FedET achieves higher performance in incremental learning compared to other baseline methods, while significantly reducing the communication cost. Specifically, compared to IDBR+FL, FedET's Enhancer structure adds a segmentation step to the regularisation and empirical replay, further improving the performance of the model.

\subsection{Computer Version(CV)}

\subsubsection{Datasets and Baselines}
We use ImageNet-Subset \cite{deng2009imagenet} and CIFAR-100 \cite{krizhevsky2009learning} to evaluate our method. We follow the same protocol as iCaRL \cite{rebuffi2017icarl} to set incremental tasks and we use the same order generated from iCaRL for a fair comparison. In detail, we compare FedET with the following baselines in the FL scenario: \textbf{iCaRL}, \textbf{BiC} \cite{wu2019large}, \textbf{PODNet} \cite{douillard2020podnet}, \textbf{SS-IL} \cite{ahn2021ss}, \textbf{DDE+ iCaRL} \cite{hu2021distilling} and \textbf{GLFC} \cite{dong2022federated}.

\subsubsection{Implementation Details}
In this CV experiment, we set $J=10$ for Enhancer Groups and $K=30$ for local. The prediction head for each group is a linear layer with a Softmax activation function. We collected samples at a store ratio of $\gamma = 0.01$. In the CIL baselines, we choose ResNet18 \cite{he2016deep} to be the backbone with cross-entropy as the classification loss. On the other hand, FedET uses frozen pre-trained ViT-Base \cite{He_2022_CVPR} as the backbone. All experiments have a batch size of 64. The training of the Enhancer used an SGD optimizer with minimum learning rate $lr_{min}=1e^{-5}$ and a base learning rate $lr_b=0.1$. In each round of global training, ten clients are randomly selected for ten local-training epochs. Selected clients are randomly given 60\% of the classes from the label space of its seen tasks.

\subsubsection{Results}
Table \ref{tab:CVcifar} and Table \ref{tab:CVImageNet} show that FedET consistently outperform all the baselines by $3.3\%\sim10.5\%$ in terms of average accuracy and reduces the communication cost to $11.0\%$ of baseline models. These results demonstrate that FedET can cooperatively train a global class-incremental model in conjunction with local clients more efficiently. Furthermore, for all incremental tasks, FedET has steady performance improvement over other methods, validating the effectiveness in addressing the forgetting  problem in FCIL.

\subsubsection{Ablation Studies}
Table \ref{tab:CVcifar} and Figure \ref{fig_ab} illustrate the results of our ablation experiments on the number of Enhancers and local clients.

+\paragraph{The number of Enhancers ($J$)} Figure \ref{fig_ab} (a) shows the model's performance in four cases, $J=1, J=5, J=10, J=15$, respectively. Compared with $J=10$, the performance of $J=1$ and $J=5$ is worse but $J=15$ is better. Since the Enhancer Select Module is frozen, in FedET, we cannot set $J = \textit{the number of classes}$. It has been verified that a suitable value of $J$ will make FedET powerful and efficient. We observed that increasing the value of $J$ makes FedET perform better at the beginning of incremental learning but degrades faster as learning progresses. We believe that the reasons may be: (1) With the increase of $J$, the requirements for the Enhancer Select Module are higher. When the  Enhancer Select Module cannot precisely perform rough classification, the model accuracy is bound to decrease. (2) As $J$ increases, the learning cost on a single Enhancer Group is lower, which means that as long as the Enhancer Select Module selects the correct group, the possibility of accurate judgment will be significantly improved.

\paragraph{The number of clients ($K$)} We tested the performance of FedET with client numbers of 30, 50, and 100, as depicted in Figure \ref{fig_ab}(b). It is clear that FedET's capacity declines as $K$ increases. The performance decrease is most noticeable when $K=100$. We believe that with an increase in $K$, the central server needs to perform distillation on more enhancers simultaneously, which causes the model to not fully converge within the specified number of iterations, resulting in a decrease in model performance.

\section{Conclusion}
FedET is an FCIL framework that can be used in many fields. Based on previous work, it introduces transformers to improve the accuracy of FCIL and increase the application field of the framework. To reduce communication costs and streamline training, only the Enhancer and its related components are designated as trainable parameters. Our experiments on datasets in NLP and CV demonstrate that FedET outperforms existing methods in FCIL while decreasing communication information by up to 90\%.


\section*{Acknowledgements}
This paper is supported by the Key Research and Development Program of Guangdong Province under grant No.2021B0101400003. Corresponding author is Jianzong Wang from Ping An Technology (Shenzhen) Co., Ltd.

\bibliographystyle{named}
\bibliography{ijcai23}

\end{document}